\documentclass{bmvc2k}
\usepackage{graphicx}   % Required for including images
\usepackage{caption}
\usepackage{tabularx}   % Required for the tabularx environment
\usepackage{xcolor}     % Required for text color
\usepackage{booktabs}
\title{Computer Vision Pipeline for Automated Antarctic Krill Analysis.}
\addauthor{Mazvydas Gudelis}{M.Gudelis@uea.ac.uk}{1}
\addauthor{Michal Mackiewicz}{M.Mackiewicz@uea.ac.uk}{1}
\addauthor{Julie Bremner}{J.Bremner@uea.ac.uk}{1,3}
\addauthor{Sophie Fielding}{sophie.fielding@uea.ac.uk}{1,2}

\addinstitution{
 University of East Anglia\\
 Norwich, UK
}
\addinstitution{
 British Antarctic Survey\\
 Cambridge, UK
}
\addinstitution{
 Centre for Environment, Fisheries and Aquaculture Science\\
 Lowestoft, UK
}
\runninghead{Gudelis, Mackiewicz, Bremner, Fielding}{CV for Antarctic Krill}

\begin{document}

\maketitle

\begin{abstract}
British Antarctic Survey (BAS) researchers launch annual expeditions to the Antarctic in order to estimate Antarctic Krill  biomass and assess the change from previous years. These comparisons provide insight into the effects of the current environment on this key component of the marine food chain. In this work we have developed tools for automating the data collection and analysis process, using web-based image annotation tools and deep learning image classification and regression models. We achieve highly accurate krill instance segmentation results with an average 77.28\% AP score, as well as separate maturity stage and length estimation of krill specimens with 62.99\% accuracy and a 1.98 mm length error respectively. 
% Our proposed pipeline correlates two views (Lateral and Dorsal) of each localised specimen and automatically classifies the length and maturity stage attributes.
\end{abstract}

%-------------------------------------------------------------------------
\section{Introduction}
\label{sec:intro}
This paper discusses the continuing research and development of computer vision tools that are to be used by British Antarctic Survey (BAS) researchers to process the Antarctic Krill image data. Antarctic Krill (Euphausia superba) are small, shrimp-like organisms that travel in dense swarms that may stretch for several kilometres and include millions of individuals. They are consumed by many marine species in the Southern Ocean, including whales, seals, and penguins, and serve a crucial role in the food chain. Krill feed on summer-blooming phytoplankton, that are reducing in number as a consequence of global warming. Antarctic Krill can adapt and are known to shrink over the winter without sufficient nourishment. Since the 1970s, krill populations have declined by 80\%, which has been linked to the melting ice that is home to the algae and plankton upon which they feed \cite{Schiermeier2010EcologistsFA, jacquet2010seafood}. In order to measure the krill biomass and evaluate the variation from year to year, annual journeys are taken to different locations around the Antarctic. %The consequences of the present climate on this crucial link in the marine food chain are shown by these comparisons. Even though scientists from BAS go to Antarctica to gather data, mistakes in capturing that data are common due to the extreme weather and sea conditions they encounter there. 

In this work we developed tools to automate the data collection and analysis of krill specimens captured in the Antarctic, using web-based image annotation tools and deep learning computer vision models. Specifically, we focus on object detection (krill specimen instance segmentation) and the automation of data processing procedures in order to transform krill image data into the form used to train deep neural networks capable of estimating two key parameters of the krill body, which are of interest to the marine biologists: its length and maturity stage.

In Section \ref{sec:bg} we cover background and related research. We further talk about the data-set we work with, methods employed, and our results in Section \ref{sec:datmet}. We conclude in Section \ref{sec:conclusions}.

% \textcolor{red}{I thought you would add machine learning results as well.}

% \textcolor{red}{Introduce subsequent sections.}

 % \textcolor{red}{I had to expand this as this gives less information on machine learning than your abstract.}

\section{Background} \label{sec:bg}
\begin{figure}
\centering
\begin{tabular}{ccc}
\includegraphics[width=3.8cm]{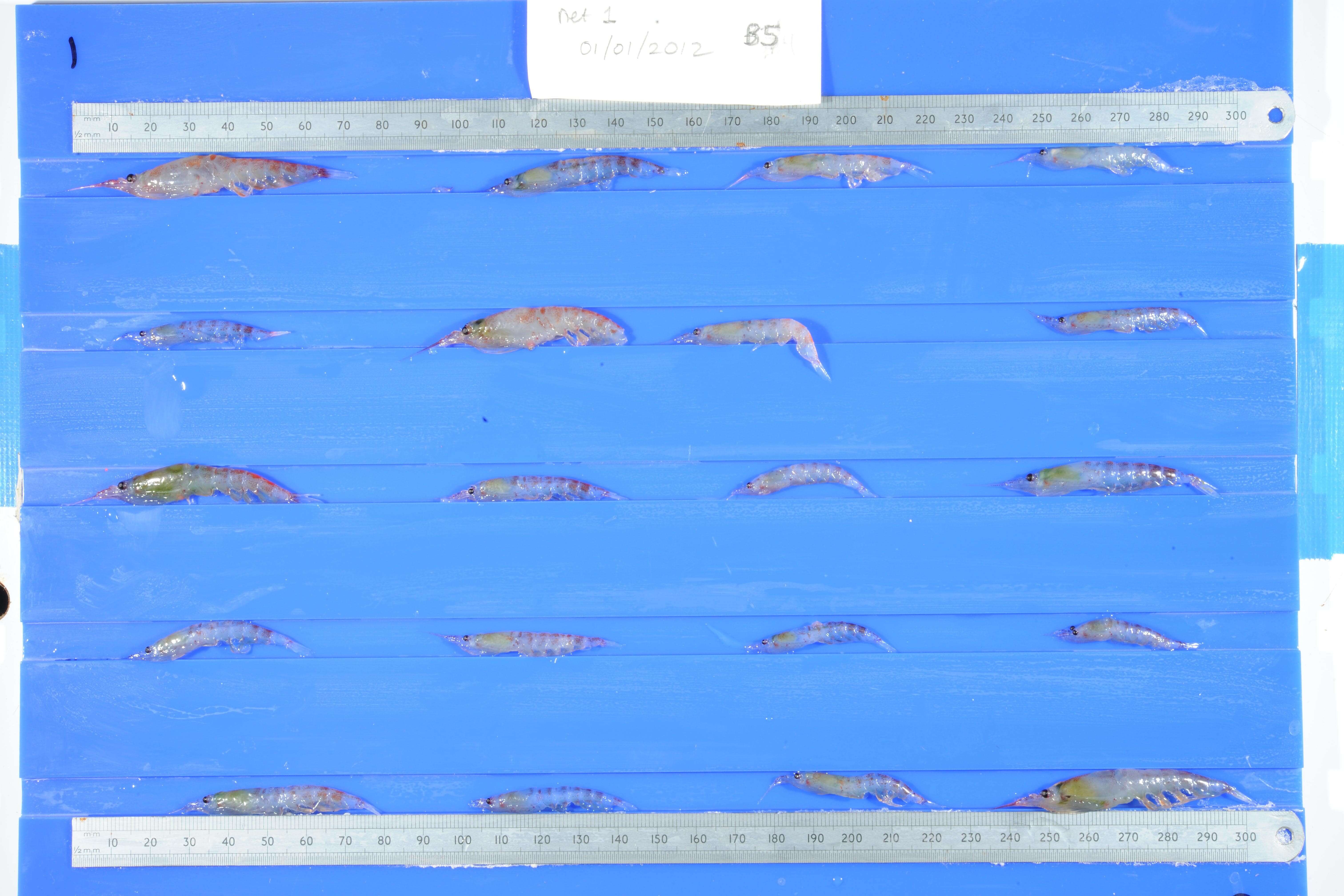} &
\includegraphics[width=3.8cm]{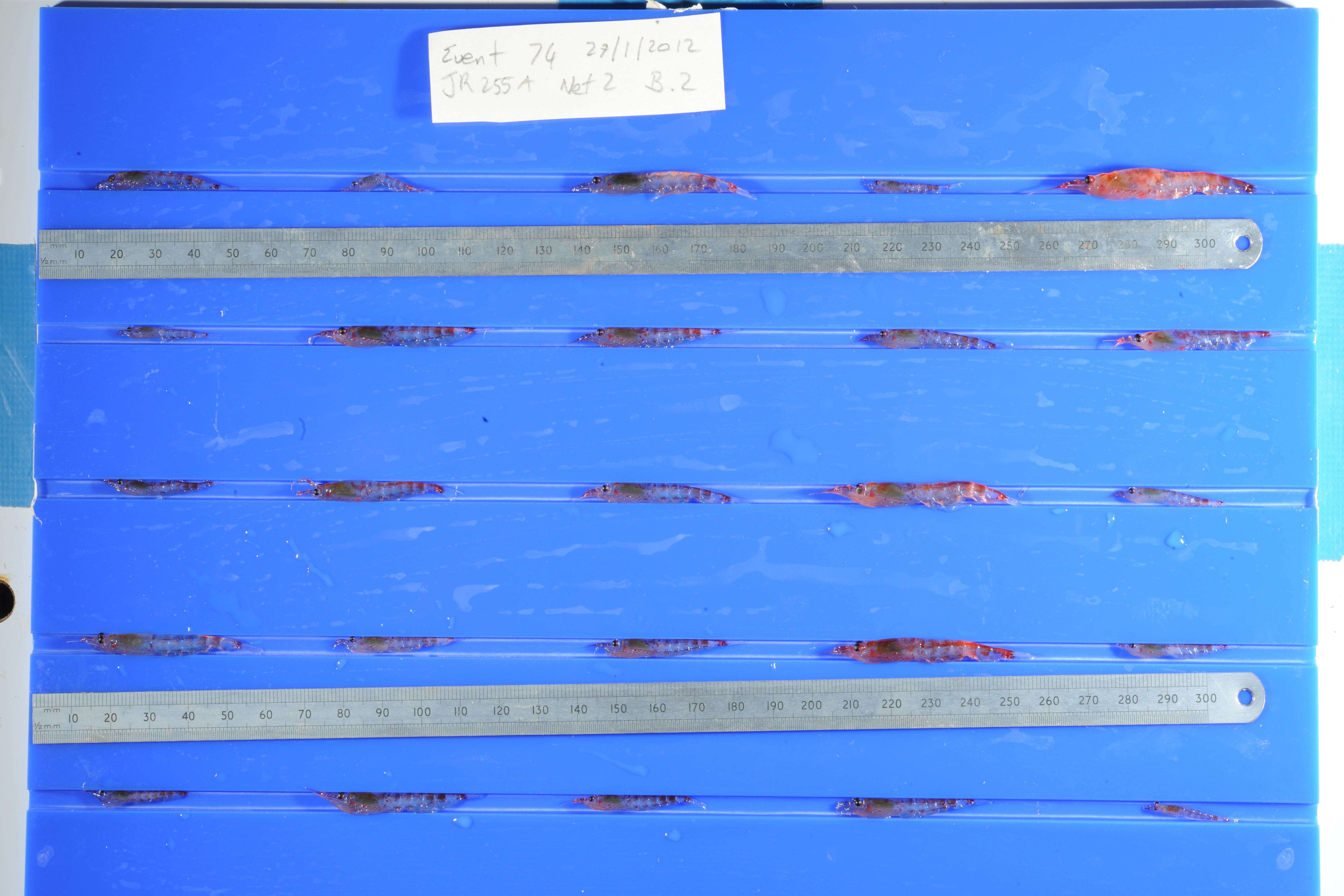} &
\includegraphics[width=3.8cm]{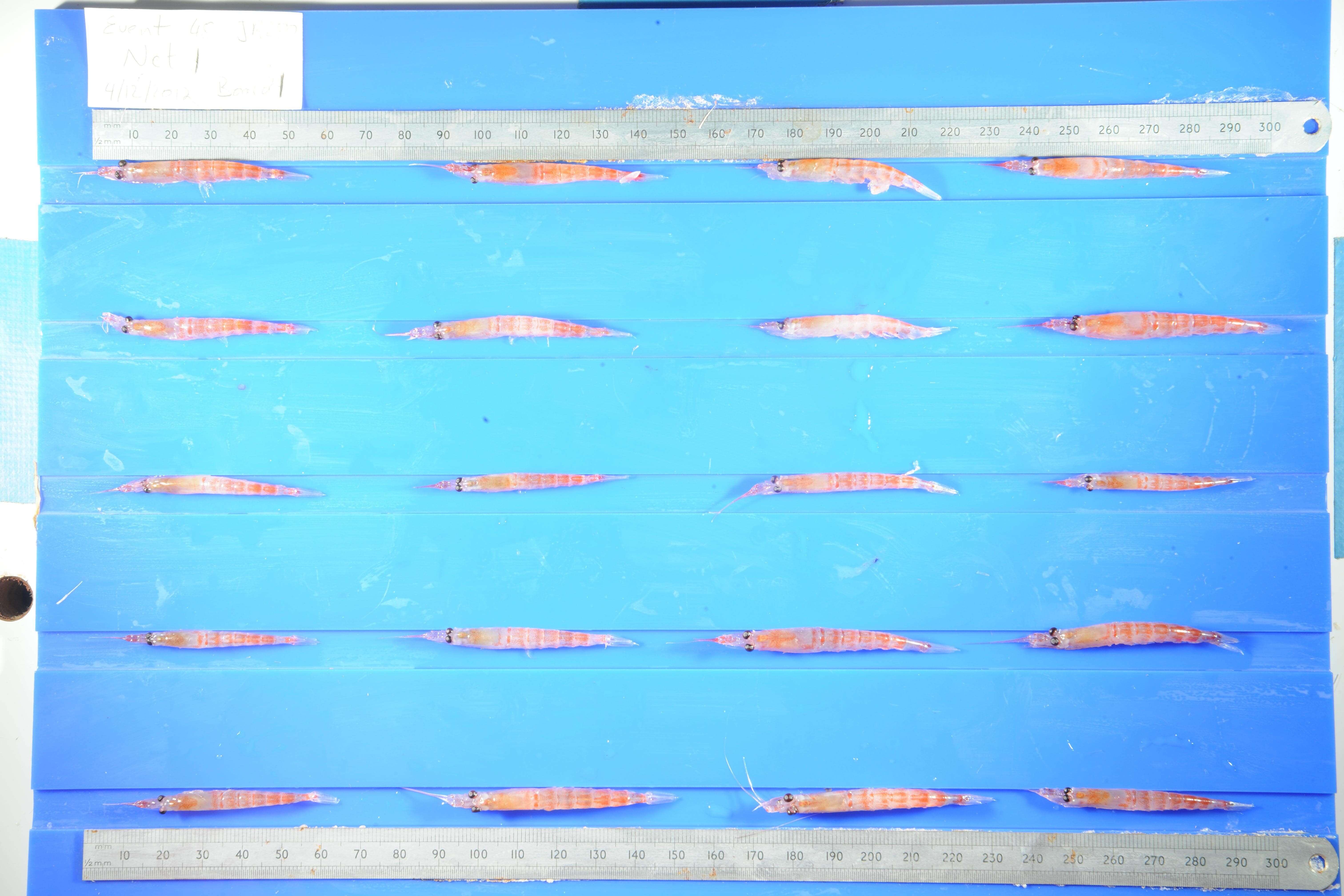} \\
(JR260B) & (JR255A) & (JR280) \\
\centering
\includegraphics[width=3.8cm]{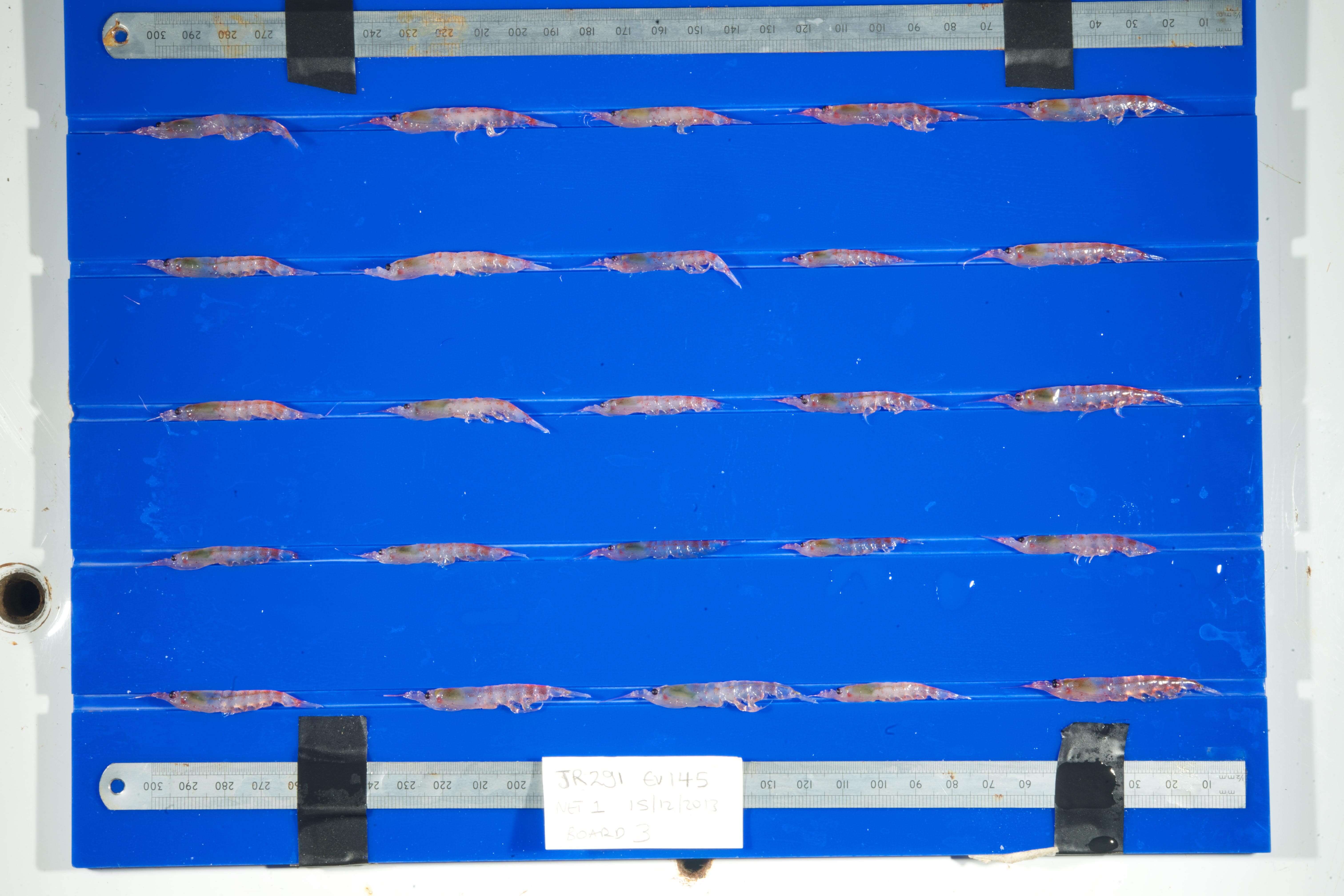} &
\includegraphics[width=3.8cm]{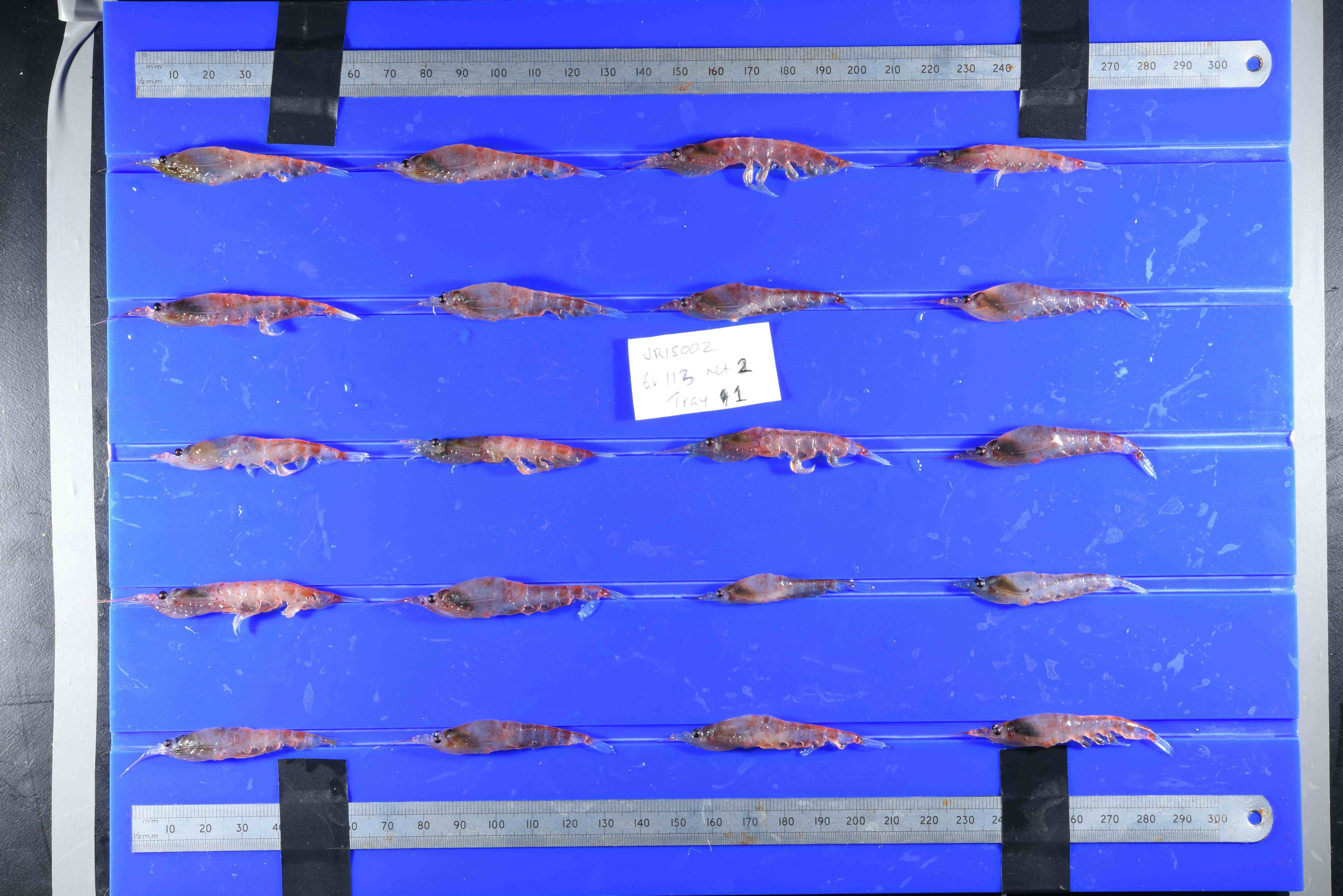} & \\
(JR291) & (JR15002) &
\end{tabular}
\vspace{1ex}
\caption{Sample of images from 5 different cruises around the Antarctic with corresponding naming encoding.}
\label{fig:teaser}
\end{figure}

Samples of Antarctic Krill are taken to evaluate the diversity of the population structure around different parts of the Antarctic and to provide parameters for the target strength model used to estimate krill biomass. Specimen length and maturity stage are the 2 main parameters that are to be inferred and both are normally assessed using long-established manual methods \citep{morris1988assessment, makarov1981stages}. The analysis is done manually for chosen specimens by physically examining their characteristics; images are only taken as digital proof of them. Taking into account the priors used to collect the data visible in Figure \ref{fig:teaser}, we suggest exploiting them to produce a consistent and user-friendly computer vision pipeline that not only exports pre-processed data for model training, but also facilitates an environment where such models can be served and used as automatic analysis tools \cite{borowiec2022deep}. 

The domain of object detection currently has a strong standpoint with many open source data-sets and models and can be utilised to localise specimens of varying size where clear object and background boundaries are visible \cite{zou2023object, szegedy2013deep, zhao2019object, redmon2016you}. Most contemporary works utilise a neural network \cite{muller1995neural} as a basis for detection and classification. Many computer vision applications already exist that solve vision problems in different domains such as agriculture \cite{bezen2020computer, fan2020line, velez2014computer}, aquaculture \cite{french2020deep}, and infrastructure \cite{quintana2015simplified}. In our case, such a system would primarily focus on krill detection and estimation of the detected specimen's maturity stage and length. 

\section{Data and Methods}
\label{sec:datmet}

The dataset we are working with consists of 457 high resolution (6048 by 4032 $px$) images containing approximately 25 krill specimens each. Photographs were taken with a Nikon DX3 and two flash guns on a stand placed approximately 1m apart, with the specimens set out on blue plastic boards (in pre-drilled grooves). The camera was set in Manual mode with an aperture of f/29 and an exposure of 1/125. The length and quality of the same krill were then evaluated physically by a marine biologist. Lengths of krill were measured according to the BAS standard, which involves measuring from the front of an animal's eye to the end of its telson and then rounding the result down to the closest millimetre \citep{morris1988assessment}. The Makarov and Denys scale \cite{makarov1981stages}, utilising the terminology established by \cite{morris1988assessment}, was used to evaluate the maturity level. The lengths and maturity level annotations will be used as the ground truth annotations in our subsequent experiments as described in Section \ref{sec:extraexp}.

In the next Section, we describe how we detect krill specimens in the aforementioned input images (Section \ref{sec:kdec}). The following Section \ref{sec:preprocessing} describes the pre-processing and data curation steps which are needed before the aforementioned image and annotation data is used in the developed web application (Section \ref{sec:web}) and in the aforementioned maturity stage classification and length regression experiments (Section \ref{sec:extraexp}).

\subsection{Krill detection}
\label{sec:kdec}

The uniform krill size and shape distribution in addition to the homogeneous environment in which they were captured provides a strong prior for object detection algorithms. The aim here is to identify each krill in an image by drawing bounding boxes or overlaying segmentation masks. 

\begin{figure}[h]
  \centering
  \captionsetup{font=small} % Set the font size for all captions in the table
  \begin{tabular}{ccc}
    \includegraphics[width=5.8cm, height=3.2cm]{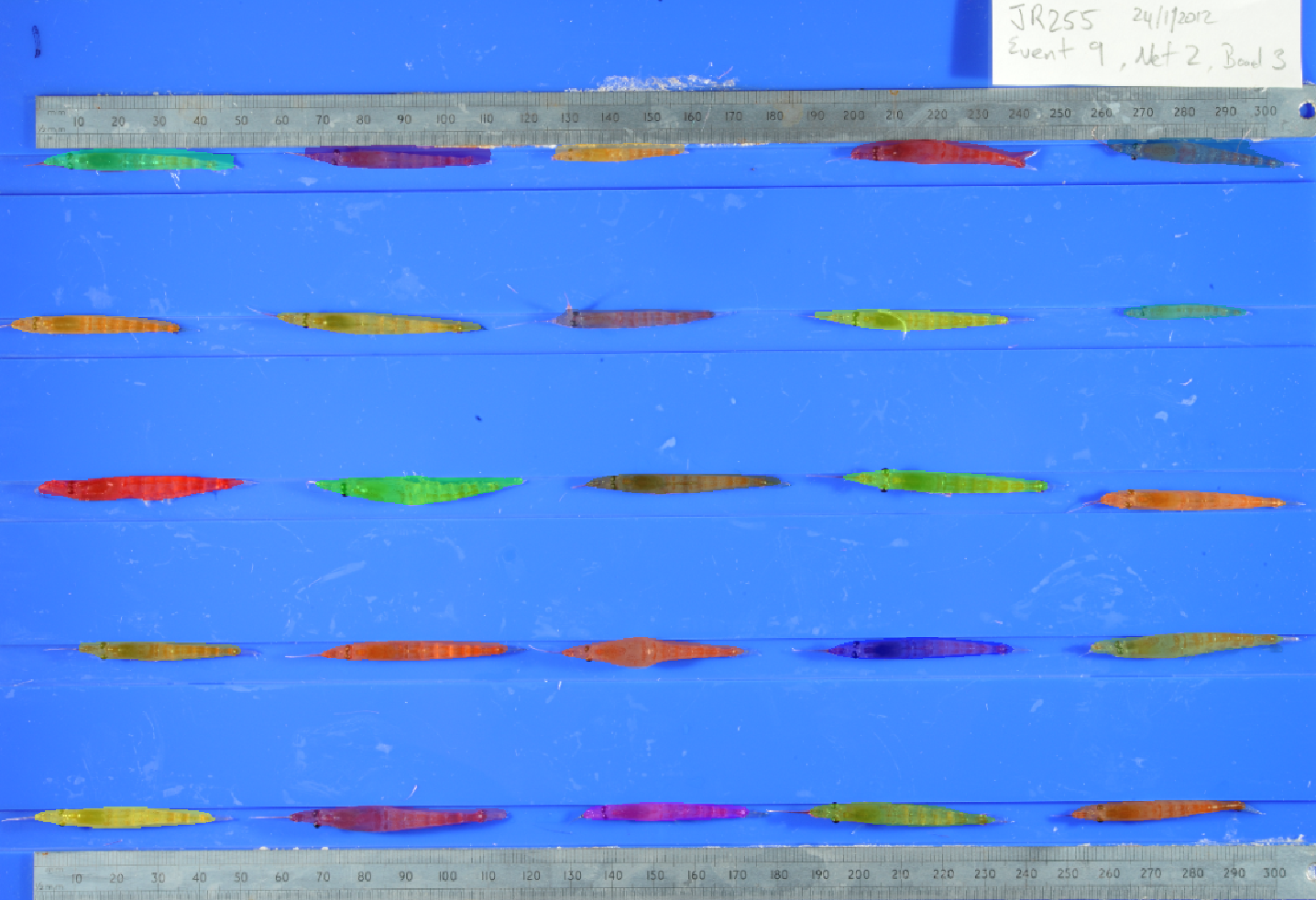} &
    \includegraphics[width=5.8cm, height=3.2cm]{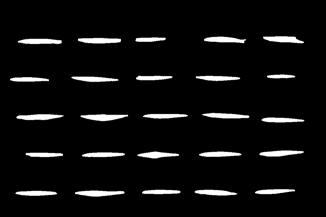}
    \\ \parbox{5.4cm}{\caption*{(a) Weakly supervised high-resolution instance labels.}} & \parbox{5.4cm}{\caption*{(b) Segmented binary image.}} \vspace{1em}\\
    \includegraphics[width=5.8cm, height=3.2cm]{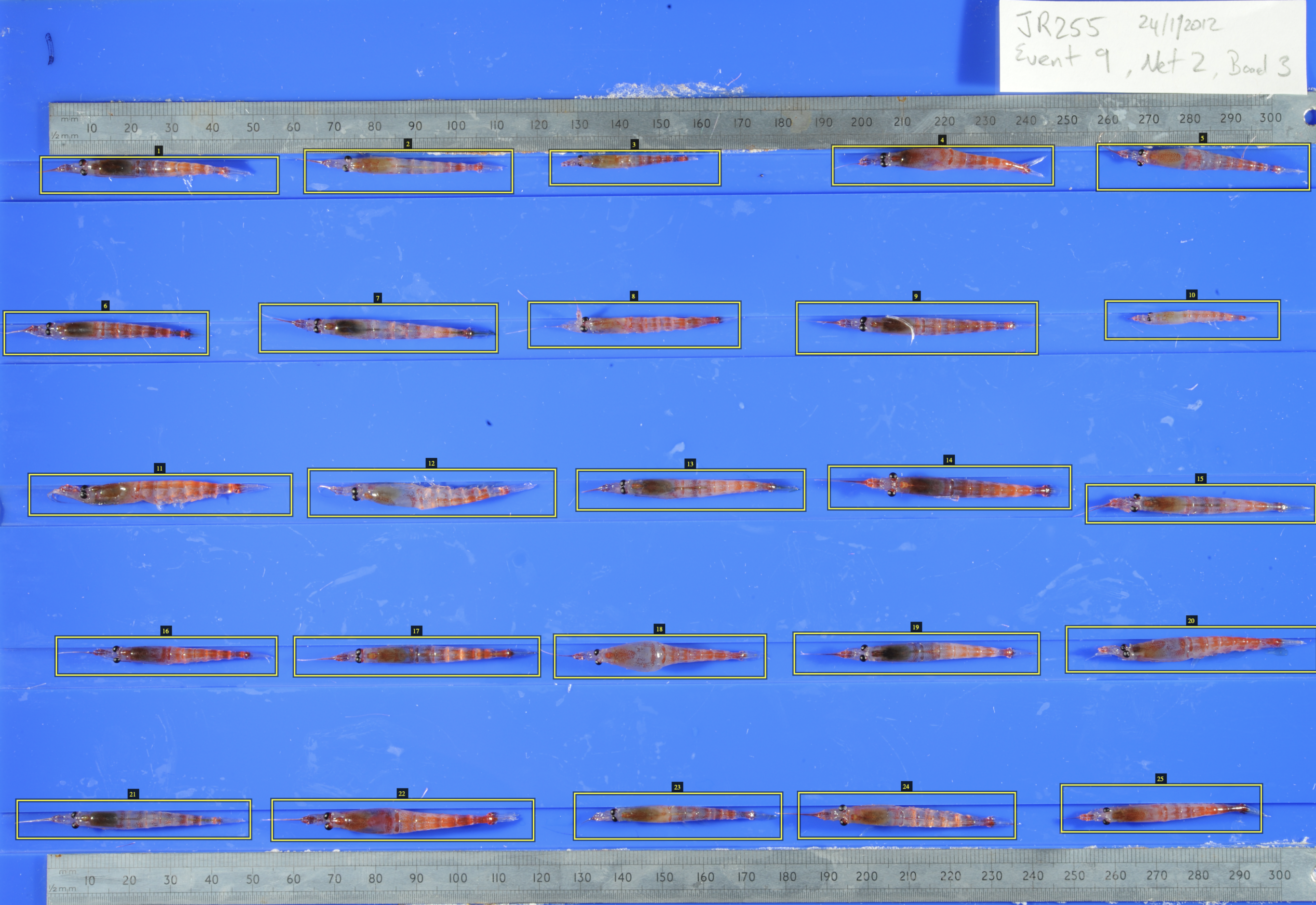} &
    \includegraphics[width=5.8cm, height=3.2cm]{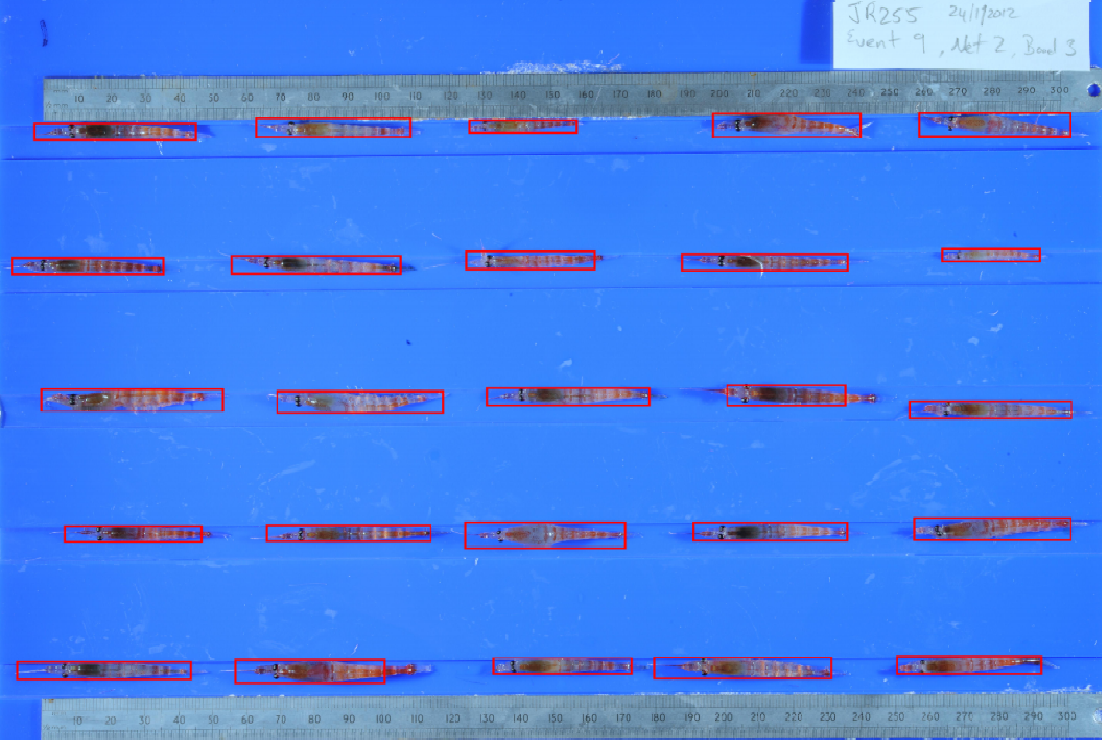} &
    \\ \parbox{5.4cm}{\caption*{(c) Interactively adjustable bounding boxes \cite{dutta2016via}.}} & \parbox{5.4cm}{\caption*{(d) Auto localised krill from segmented image.}} \vspace{1em}\\
    \includegraphics[width=5.8cm, height=3.2cm]{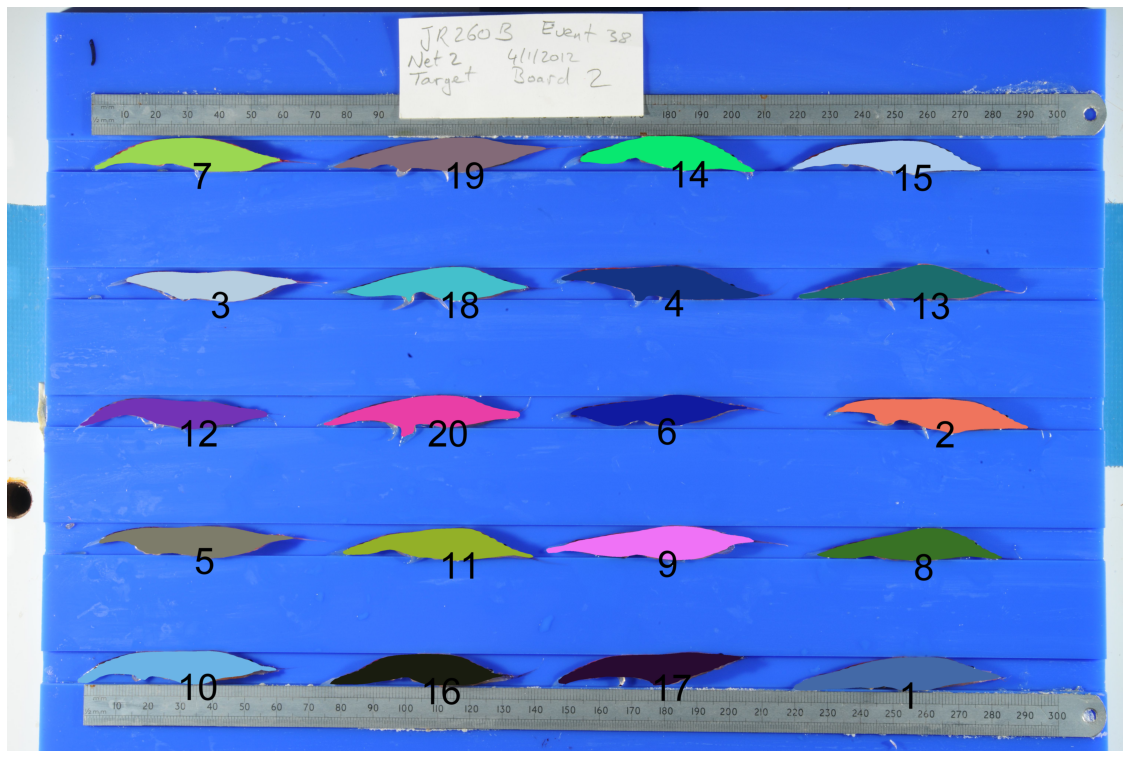} &
    \includegraphics[width=5.8cm, height=3.2cm]{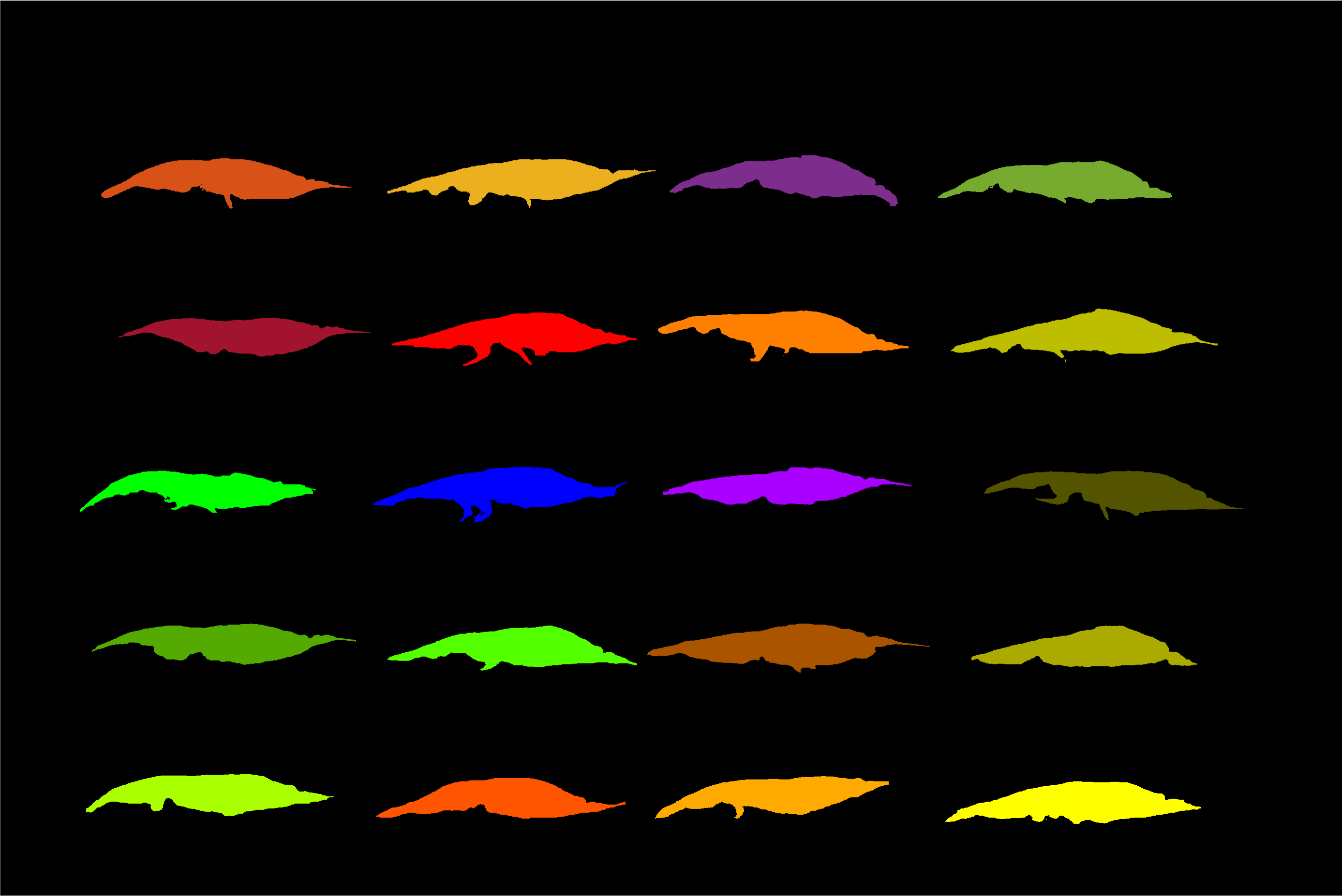} 
    \\ \parbox{5.4cm}{\caption*{(e) Segmentation prediction visualisation for test set.}} & \parbox{5.4cm}{\caption*{(f) Segmentation ground truth visualisation of previous image.}} \\
  \end{tabular}
  \vspace{2em}
  \caption{Krill detection image processing steps. The proposed detection pipeline produces bounding box coordinates as well as instance masks. Bounding boxes can be adjusted through VIA \cite{dutta2016via}.}
  \label{fig7}
\end{figure}

\begin{figure} 
  \centering
  \begin{tabular}{cccc}
    \includegraphics[width=5.5cm, height=3.5cm]{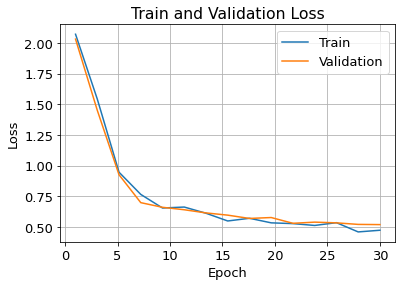} &
    \includegraphics[width=5.5cm, height=3.5cm]{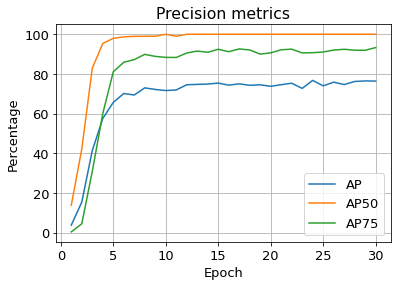} \\
    (a) & (b) \\
    \includegraphics[width=5.5cm, height=3.5cm]{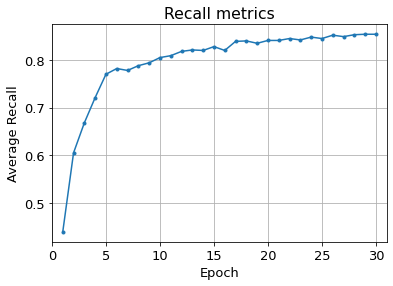} &
    \begin{minipage}[c]{4.8cm}
    \vspace{-3.7cm}
      \centering
      {\fontsize{7}{9}\selectfont
      \begin{tabular}{cccc}
        {\color[HTML]{000000} \textbf{CRUISE}} & {\color[HTML]{000000} \textbf{AP}} & {\color[HTML]{000000} \textbf{AP50}} & {\color[HTML]{000000} \textbf{AP75}} \\
        \hline
        {\color[HTML]{000000} JR255A}        & 79.32 \%                                          & 98.97 \%                             & 93.71 \%                             \\
        {\color[HTML]{000000} JR260B}        & 81.17 \%                                          & 98.99 \%                             & 96.99 \%                             \\
        {\color[HTML]{000000} JR280}         & 74.37 \%                                          & 100 \%                               & 94.62 \%                             \\
        {\color[HTML]{000000} JR291}         & 79.22 \%                                          & 100 \%                               & 96.58 \%                             \\
        {\color[HTML]{000000} JR15002}       & 72.32 \%                                          & 98.99 \%                             & 76.89 \%                             \\
        \hline
        {\color[HTML]{000000} \textbf{AVG}}  & 77.28 \%                                          & 99.39 \%                             & 91.71 \%                            
      \end{tabular}
      }
    \end{minipage}
    \\ (c) & (d) 
  \end{tabular}
  \vspace{1em}
  \caption{Model convergence plot (a). Mask precision metrics for test set (b). Mask recall metrics for test set (c). Table showing mask precision for cross validation sets (d).}
  \label{detect} 
\end{figure}

The krill detection pipeline uses a Mask R-CNN \citep{rick} to detect each specimen's outline. We opt for an approach that ensures high resolution instance masks along with bounding box coordinates of the specimens are obtained. To do this in an effective way, weakly supervised method from bounding boxes \cite{tian2021boxinst} is used. We manually label all 457 full board krill images with bounding box labels and produce high-resolution instance masks visible in Figure \ref{fig7} (a) and (b). Each specimen's mask within the segmentation map is decoded to retrieve the bounding box coordinates. This acts as an automatic krill detector when passed to the labelling tool (see Figure \ref{fig7} (d)) that we based on the VIA \cite{dutta2016via}. The bounding box parameters can be adjusted manually by the user as shown in Figure \ref{fig7} (c), allowing for more specific user needs. The fully supervised krill segmentation algorithm assumes a varying data-set $D=\left \{ (x_{i},y_{i})|1\leq i\leq N \right \}$ where $x_{i}$ are the inputs, $y_{i}$ are the ground truth masks, $\mathcal{L}$ is the loss, and the training/testing procedure is denoted as $D_{TR}\cup D_{TE}$. The goal of the learning framework is to learn $f(x)=y$ by minimising $\sum _{}(x_{i},y_{i})\in D_{TR}$ $\mathcal{L}(f(x_{i}),y_{i})$ in the hope of generalising to $\sum _{}(x_{i},y_{i})\in D_{TE}$ $\mathcal{L}(f(x_{i}),y_{i})$. $D_{TR}$ and $D_{TE}$ are split with a random 80\% to 20\% ratio. A pre-trained ResNet50-FPN Mask R-CNN variation \citep{rick} model is fine-tuned on $D_{TR}$ and tested on $D_{TE}$. In Figure \ref{detect}, we show average precision and recall metrics for the experiment.

We further conduct 5 leave-one-cruise-out experiments \citep{Arlot_2010}. In each experiment \( i \), a unique cruise \( c_i \in \mathcal{C} \) is designated for testing, while the remaining cruises are utilised for training. The results for krill detection can be seen in Figure \ref{detect}. Our proposed Krill Tool which we will discuss in Section \ref{sec:web} utilises the model at a click of a button, and generates global instance masks for every specimen visible in an image. Figure \ref{fig7} (e), (f) show mask predictions as compared to the labels for krill images from the test set. In total, this approach outputs multiple image areas of interest (krill instances) in bounding box and pixel mask format.

\subsection{Data pre-processing}
\label{sec:preprocessing}

\begin{figure}
  \centering
  \begin{tabular}{cc}
    \includegraphics[width=5.8cm, height=3.2cm]{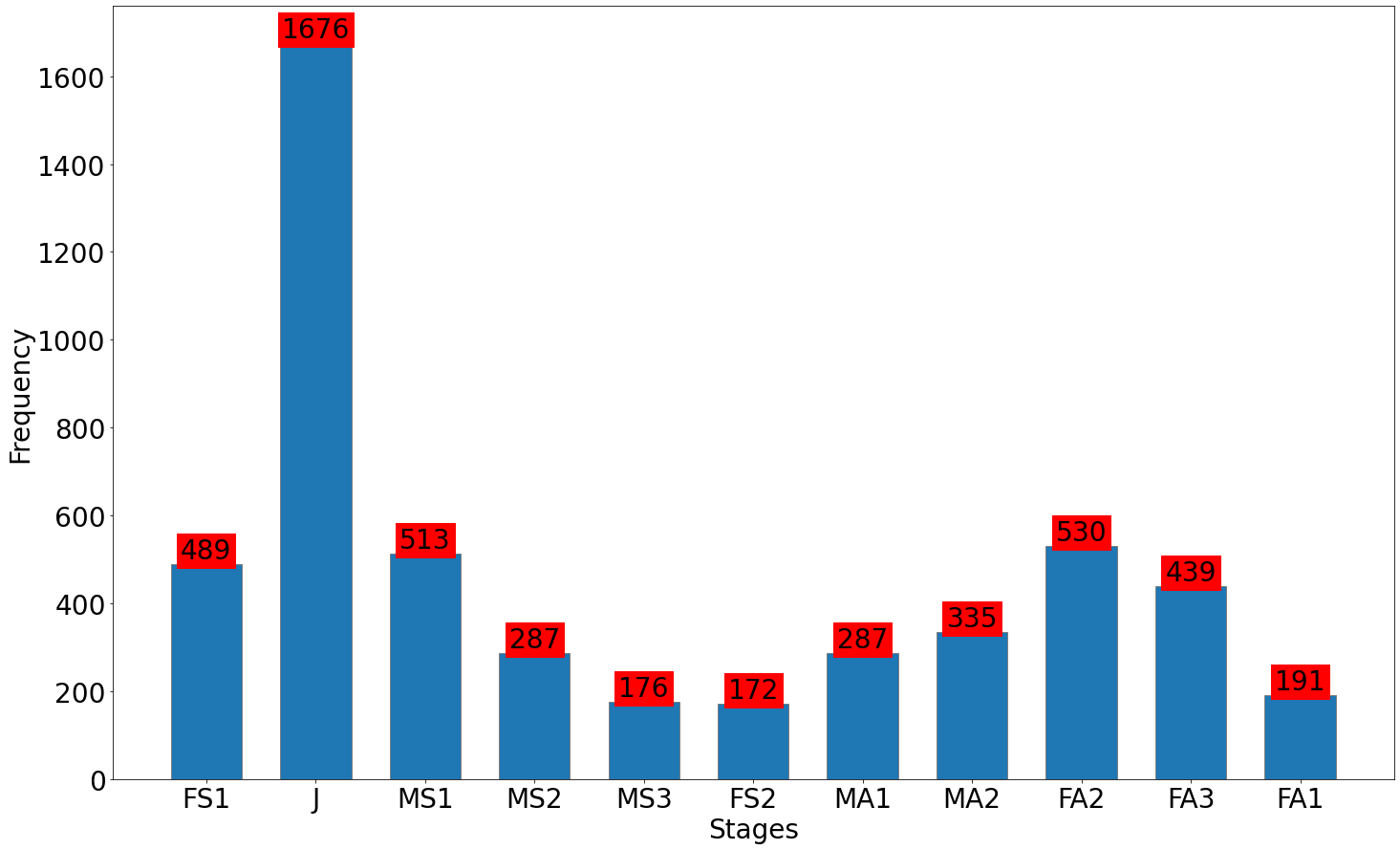} &
    \includegraphics[width=5.8cm, height=3.2cm]{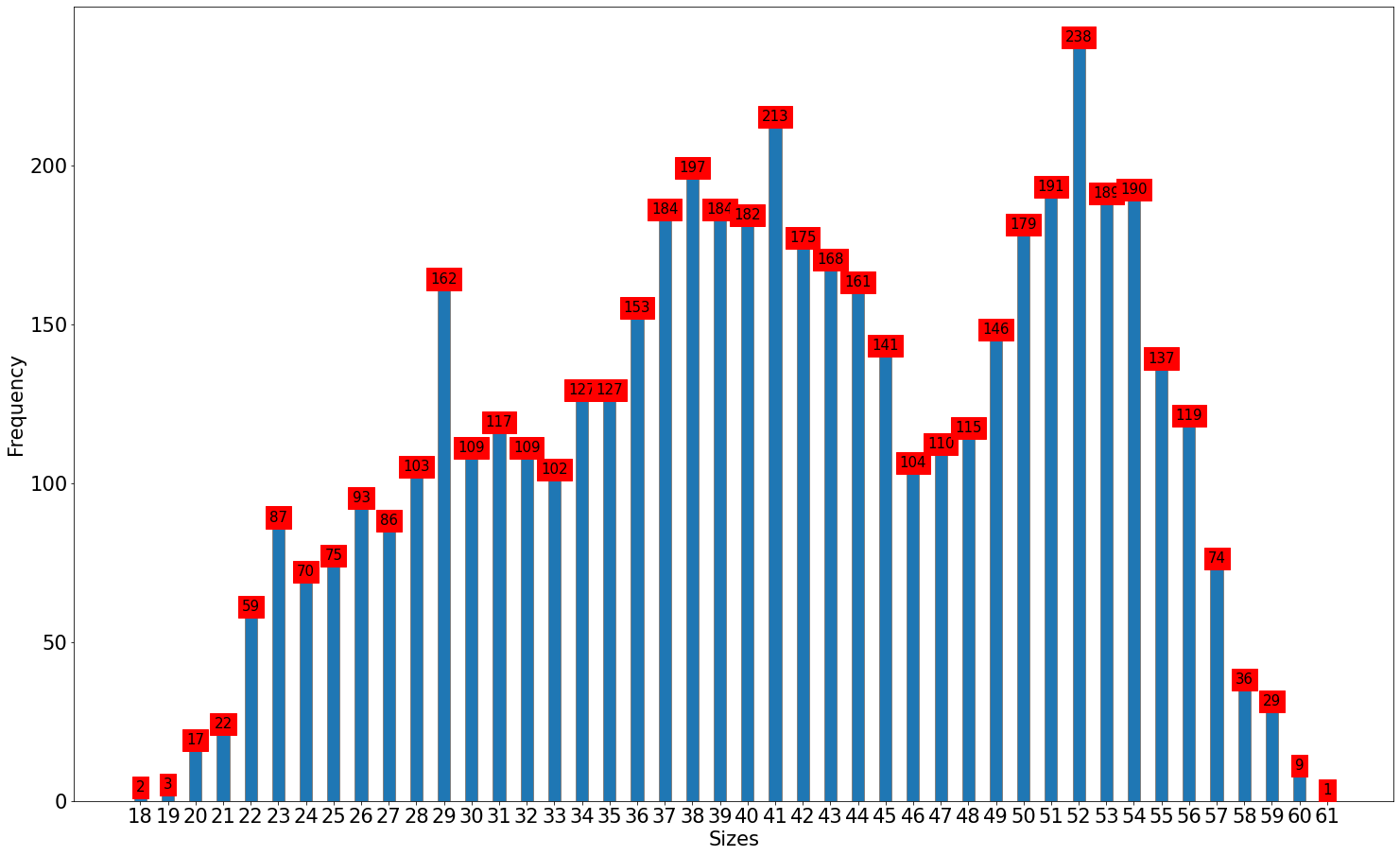} \\
    (a) & (b) \\
  \end{tabular}
  \vspace{1em}
  \caption{Krill maturity stage distribution across 11 stages (a). Body measurement distribution in millimetres (b).}
  \label{visvis} 
\end{figure}

The data preparation process starts with the assignment of two essential target parameters to each specimen (length and maturity stage). As indicated at the beginning of Section \ref{sec:datmet}, BAS provides ground truth data gathered by a field expert for the vast majority of specimens which is delivered in the MS Excel spreadsheet format. The given spreadsheet needs to be processed to a format usable by the subsequent downstream tasks. This includes enforcing naming conventions and dataset consistency. The sample of the pre-processed data can be seen in Table \ref{tab:my-text database export}. This assumes an initial collection of all data and demonstrates that each view is regarded to have the same data attributes. The $ID$ columns, which are also the picture file names, provide access to the related photos. The first two columns include the two attributes of interest: \textit{length}, which was measured in millimetres, and a \textit{maturity stage} indicator, as described in \cite{makarov1981stages}. The two $ID$ columns are formed by concatenating the cruise, image sequence number, file name and index, which is counted top-left to bottom-right as shown in Figure \ref{fig7} (c). Catch specific details such as event, net, and board numbers are also recorded. The table is manually checked for clearly visible errors such as incorrect view or parameter assignment. The $x$, $y$, $width$, $height$ are used with the image name to get the localised specimen views as shown in Figure \ref{webapp} (b). The bounding box dimensions of localised krill specimens have a mean of $951$ x $256$ pixels and standard deviation of $222$ x $62$ pixels. The largest and smallest recorded samples have the dimensions of $1663$ x $488$ and $312$ x $47$ respectively. From this image pre-processing stage the 10524 entries mentioned in Table \ref{tab:my-text database export} are extracted from the high-resolution full-board images for further processing. The extraction process is done for the RGB images as well as their corresponding Mask R-CNN binary masks. Figure \ref{webapp} (b) shows the 11 selected maturity stages that are analysed further. Only krill images that contain both lateral and dorsal views and all key parameters are used, meaning after data cleaning we accumulate 5095 labelled unique krill specimens spanning across 10190 images. The sample distribution with regards to the 2 key parameters can be seen in Figure \ref{visvis}. Both plots clearly indicate an imbalanced data-set, meaning further steps must ensure the reduction of bias as discussed in \cite{buda2018systematic}. To enforce a constant spatial feature lock each image is padded up to $1700$ x $500$ pixels as this resolution falls slightly above the maximum recorded one. Each krill specimen region is placed at the centre of the padded image which uses colour of $RGB=[56, 127, 245]$ as background. The samples that included incorrect maturity labels such as M1, A2, U and classes that had $\le$100 samples (FA5, FS3, MA3) were excluded during the data pre-processing steps.

\begin{table*}
\resizebox{\textwidth}{!}{
\begin{tabular}{@{}llllllllllllll@{}}
\toprule
\textbf{length} & \textbf{maturity} & \textbf{cruise} & \textbf{x} & \textbf{y} & \textbf{width} & \textbf{height} & \textbf{ID} & \textbf{Alternative view ID} & \textbf{position} & \textbf{event} & \textbf{net} & \textbf{board} \\ \midrule
34 & FS1 & JR255A & 469 & 751 & 869 & 114 & JR255A\_krill\_image\_73.jpeg-1 & JR255A\_krill\_image\_74.jpeg-1 & Dorsal & 78 & 2 & 3 \\
23 & J & JR255A & 1368 & 869 & 537 & 118 & JR255A\_krill\_image\_73.jpeg-2 & JR255A\_krill\_image\_74.jpeg-2 & Dorsal & 78 & 2 & 3 \\
25 & J & JR255A & 2207 & 851 & 560 & 123 & JR255A\_krill\_image\_73.jpeg-3 & JR255A\_krill\_image\_74.jpeg-3 & Dorsal & 78 & 2 & 3 \\
29 & J & JR255A & 3172 & 819 & 746 & 168 & JR255A\_krill\_image\_73.jpeg-4 & JR255A\_krill\_image\_74.jpeg-4 & Dorsal & 78 & 2 & 3 \\
40 & MS1 & JR255A & 4319 & 783 & 1038 & 191 & JR255A\_krill\_image\_73.jpeg-5 & JR255A\_krill\_image\_74.jpeg-5 & Dorsal & 78 & 2 & 3 \\ \bottomrule
\end{tabular}%
}
\newline
\vspace{1em}
\caption{Krill tool text database export (first 5 sample entries). Each entry is a registered krill view with corresponding parameters. Total 10524 entries.}
\label{tab:my-text database export}
\end{table*}

\smallbreak

\subsection{Web application}
\label{sec:web}
\begin{figure}
  \centering
  \begin{tabular}{cc}
    \includegraphics[width=6.7cm, height=4.2cm]{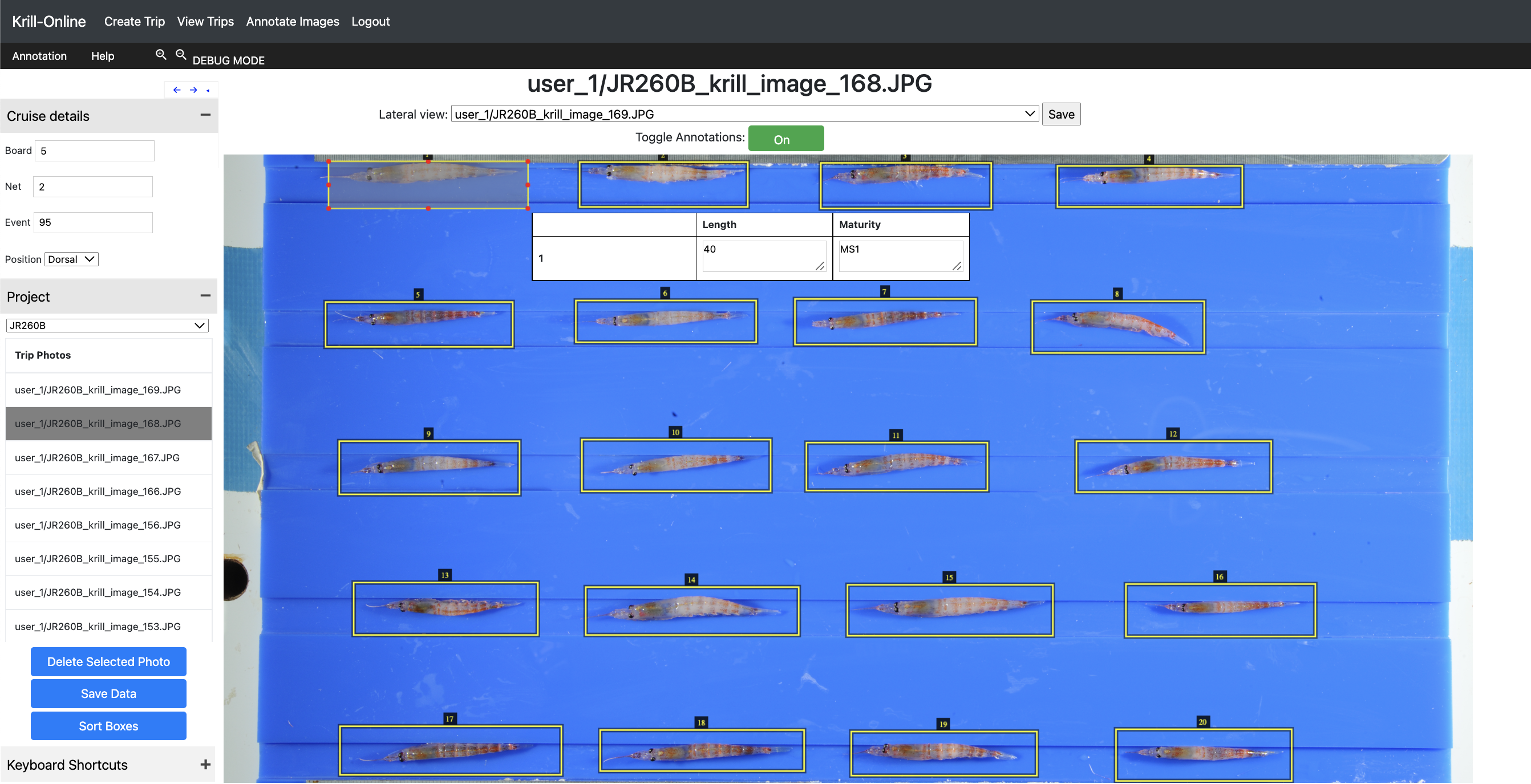} &
    \includegraphics[width=5.5cm, height=5cm]{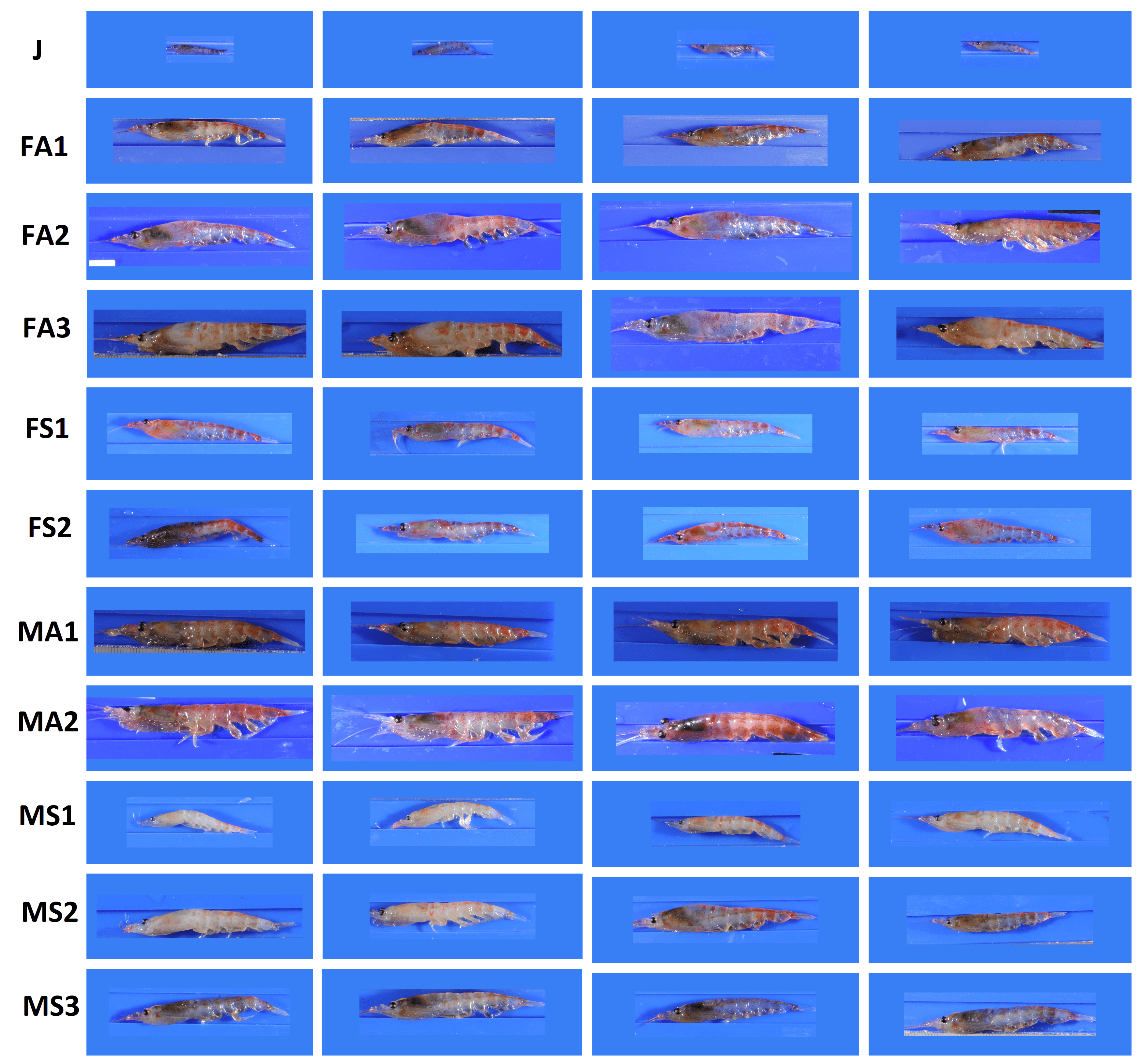} \\
    (a) & (b) \\
  \end{tabular}
  \vspace{1em}
  \caption{Web application user interface image labelling view (a). Pre-processed krill images (b).}
  \label{webapp} 
\end{figure}

A web-based application, named Krill Tool, has been developed using mainly Python Django \citep{django} and VGG image annotator \citep{dutta2016via} to simplify the data annotation and pre-processing procedures. The application enables researchers to transform collected data into a format suitable for classifier training. The application interface provides an annotation system that enables researchers to rapidly assign parameters to each specimen. Each full-board image uploaded is allocated an alternative view, which displays either the Lateral or Dorsal view of each specimen. The system requires the user to associate each image view with its alternative. The application is intended for the manual entry of user data in order to designate the length and maturity labels for each specimen. It is anticipated that this system will facilitate future expansion of ground-truth images and annotations. Figure \ref{webapp} (a) demonstrates the layout of the labelling interface.

\subsection{Maturity classification and length regression}
\label{sec:extraexp}

The 10190 extracted and pre-processed images seen in Figure \ref{webapp} (b) are used to separately train a maturity stage classifier and a length regressor. Current experiments only use the bounding-box produced images and not the mask. For consistency, the same ResNet50 backbone \cite{he2016deep} is used as in Section \ref{sec:kdec}. We treat Lateral and Dorsal as separate data-sets and reduce resolution from the maximum one down to 340x100 over several steps. As shown in Figure \ref{visvis}, the collected data-set is unbalanced and needs certain approaches to account for that \cite{buda2018systematic}. Hence, the following regression and classification experiments were executed where class weights were calculated $w_{j}=\frac{n}{s_{j}\cdot s}$, where $w_{j}$ is the weight for each class, $n$ the total number of samples in the data-set, $s$ the total number of unique classes, $s_{j}$ the total number of samples of the respective class. Parameter estimation experiments use a similar 80\%/20\% train/test split, although the training pipeline during these experiments is slightly altered - we allow the whole model to backpropagate, use more colour based augmentations, reduce the learning rate to \( 1 \times 10^{-3} \) and double the training time to 60 epochs. 

\begin{table}[h]
\centering
\begin{tabular}{lcc|cc}
\toprule
Resolution & \multicolumn{2}{c}{Lateral} & \multicolumn{2}{c}{Dorsal} \\
\hline
 & Length & Maturity & Length & Maturity \\
\midrule
340x100 & 2.23 & 58.41 & 3.12 & 57.63 \\
680x200 & 2.26 & 59.85 & 2.33 & 57.09 \\
1020x300 & 2.13 & 58.66 & 2.28 & 60.00 \\
1360x400 & 2.42 & 62.99 & 2.19 & 57.09 \\
1700x500 & 1.98 & 61.65 & 2.04 & 58.16 \\
\bottomrule
\end{tabular}
\vspace{1em}
\caption{ResNet50 Results for Length regression error in mm (lower is better) and Maturity classification accuracy (higher is better)}
\label{combined_table}
\end{table}

The results can be seen in Table \ref{combined_table}. Overall, the Lateral view provides slightly better results overall, which is logical given that physical analysis of length and morphology has been carried out on the Lateral view by the BAS researchers.

As expected, higher image resolutions generally yield lower length regression errors as well as higher maturity stage classification accuracies. While the above trend is clear, the results also show that higher image resolutions do not always achieve the higher classification accuracy as can be seen from the maturity column for the Dorsal view images. 

\begin{figure}[h]
  \centering
  \begin{tabular}{cc}
    \includegraphics[width=6cm]{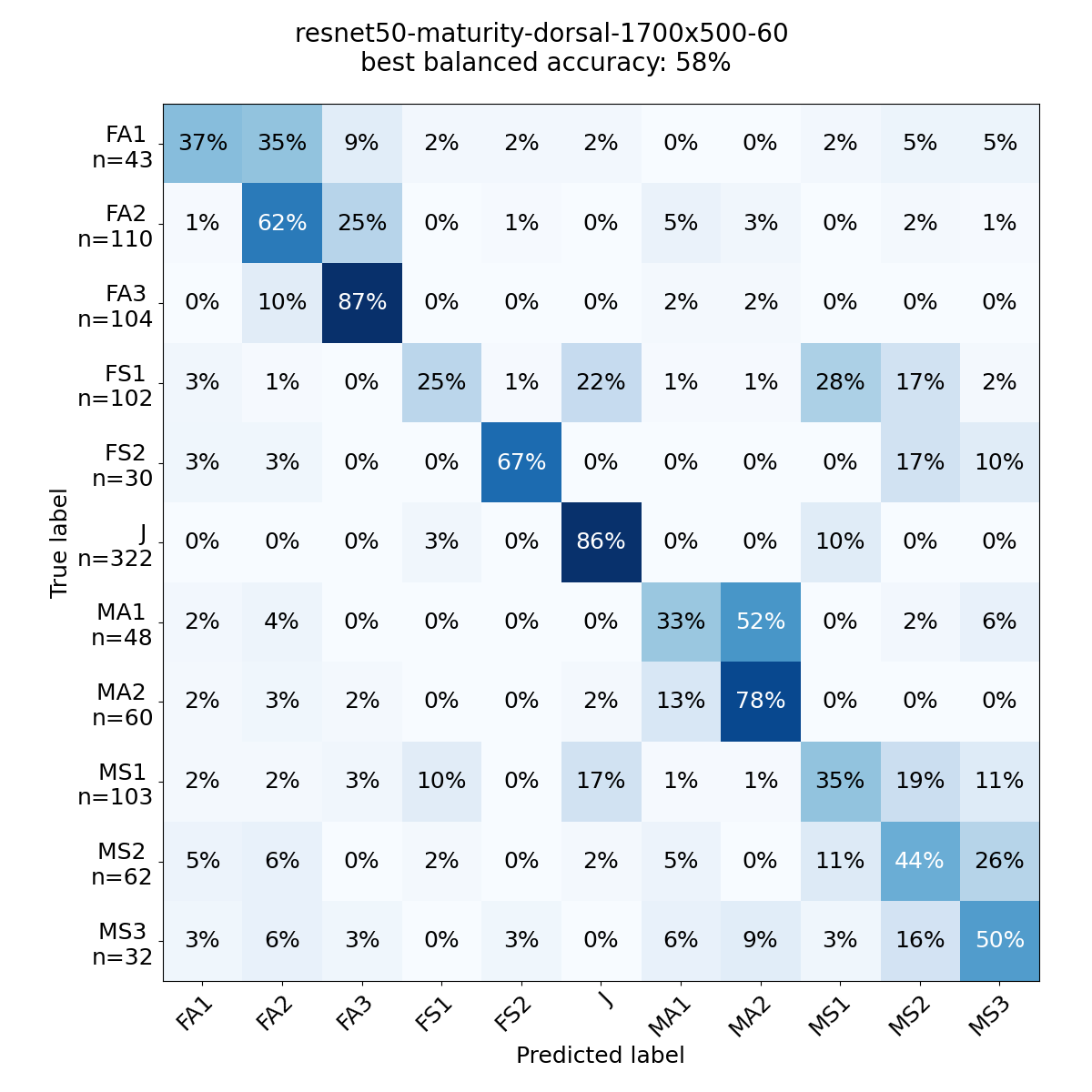} &
    \includegraphics[width=6cm]{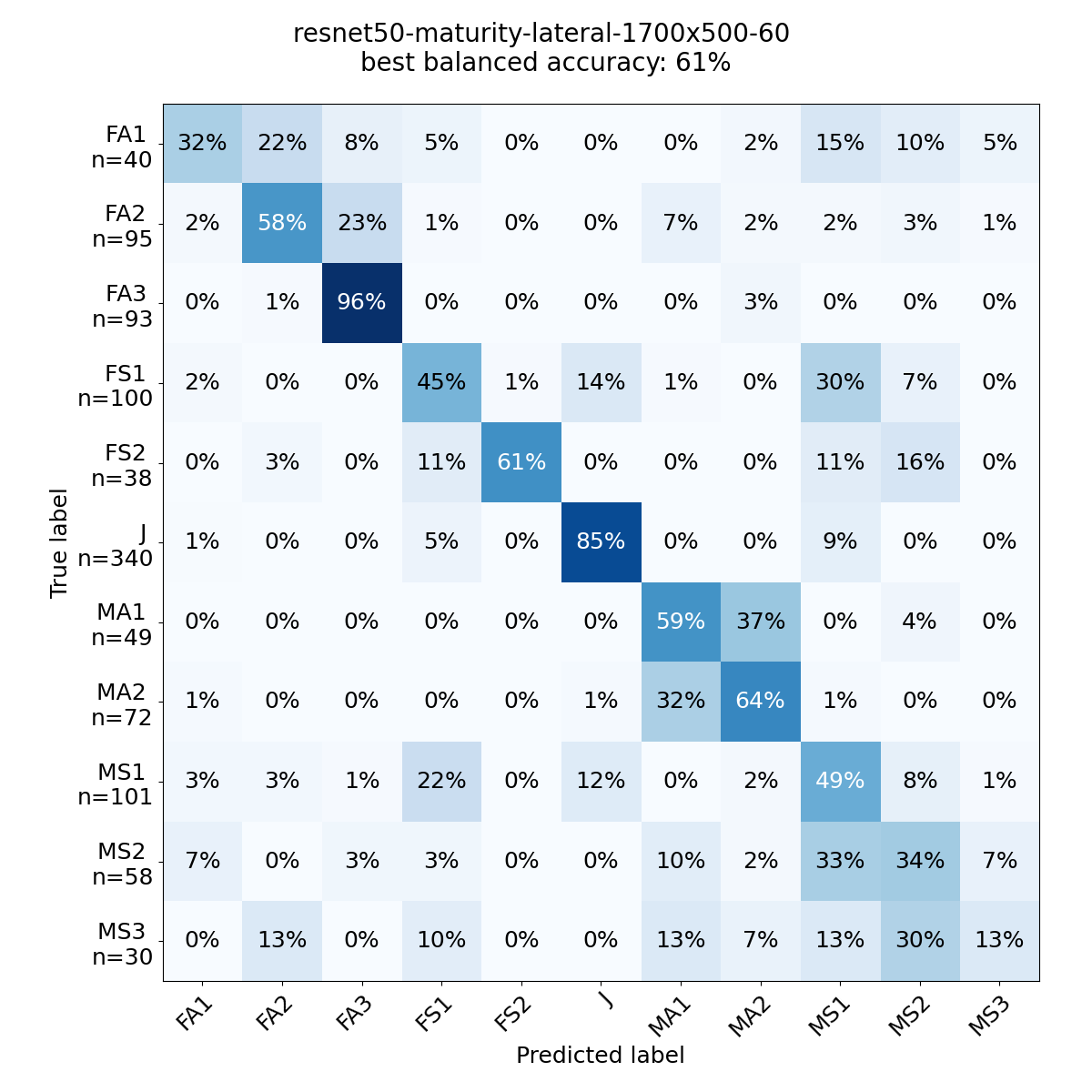} \\
    (a) & (b) \\
  \end{tabular}
  \vspace{1em}
  \caption{Dorsal model, 1700x500 pixel images (a). Lateral model, 1700x500 px images (b).}
  \label{confma} 
\end{figure}

The two confusion matrices shown in Figure \ref{confma} give a detailed visualisation of the maturity stage classifier's predictions. Similar confusion matrix patterns are visible across both dataset views. The junior maturity stage (class J) is the most predominant one and also achieves the highest classification accuracies in every experiment due to its distinct differences from other classes. Other misclassifications follow a predictable patter - there are some misclassfications between MS1 and FS1 maturity stages, as well as between consecutive maturity stages e.g. MA1 and MA2; and MS1, MS2 and MS3. We need to note that our ground truth annotations have been provided by a single marine biologist and given the difficulty of establishing krill maturity stage manually, there is a certain level of uncertainty related to the ground truth annotations we used in the above experiments. This level of uncertainty cannot be qualified without additional human annotations from other marine biologists \cite{rs12122026}, which would be of course helpful from the point of view of assessing the performance of the automated system in the context of human inter-observer error. As the current data was gathered by one marine biologist on board a research vessel, we are unable in this case to provide any inter-observer metrics.

\begin{figure}[h]
  \centering
  \begin{tabular}{cc}
    \includegraphics[width=6cm]{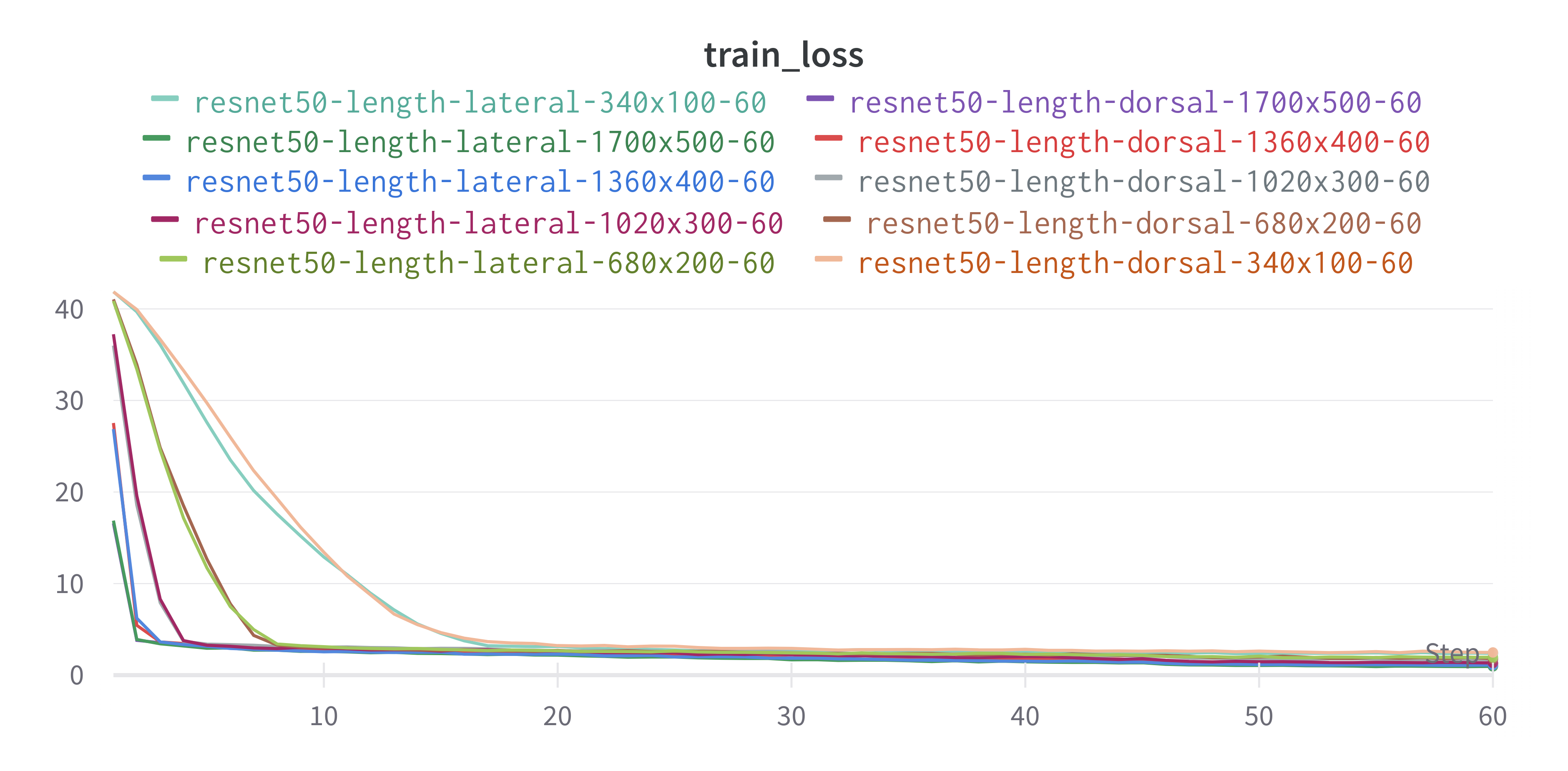} &
    \includegraphics[width=6cm]{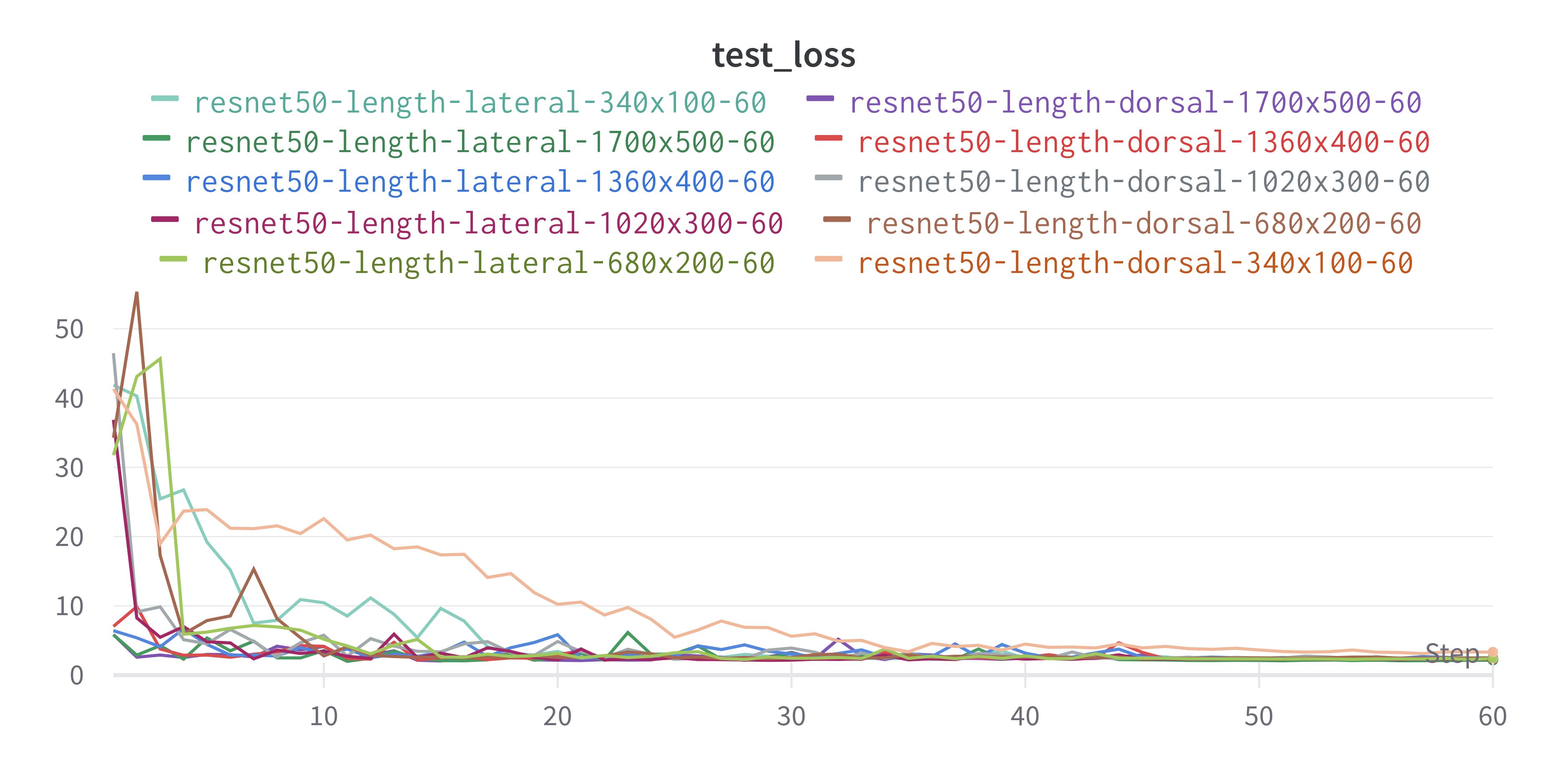} \\
    (a) & (b) \\
  \end{tabular}
  \vspace{1em}
  \caption{Train loss (RMSE) convergence (a). Test loss (RMSE) convergence (b).}
  \label{losses} 
\end{figure}

The length regressor proposed in this study employs the root mean square error (RMSE) loss function  \cite{hodson2022root}. Consequently, the length regression errors observed in Table \ref{combined_table} and Figure \ref{losses} represent the average millimetric deviation of the predictions. Convergence is reached within a few epochs for higher resolution models. Lower resolution models take longer to converge, but their performance is on par with higher resolution models. The Lateral view on average yields a 2.20 mm error, whereas the Dorsal one, as expected, is slightly higher at 2.39 mm.

\section{Conclusions and Future Work}\label{sec:conclusions}
By leveraging the power of computer vision and deep learning in marine biology, here we demonstrate the capability of computer vision tools to automate the collection and analysis of Antarctic Krill image data. The essential function of Antarctic Krill in the marine food chain and the unprecedented challenges posed by climate change necessitate an accurate population assessment of krill. 
% \textcolor{red}{Experiments utilising high-resolution instance masks demonstrate increased sensitivity in krill parameter estimation. - MM: I would drop this sentence}  

The major product from this work is the Krill Tool, a web-based application that facilitates the data annotation and pre-processing to ensure consistency and efficacy in data preparation. The application provides a user-friendly pipeline that replaces manual methods with an automated framework, with promising initial results in maturity classification and length estimation. We achieve a highest accuracy of 62.99\% on the Lateral view for maturity classification as well as a 1.98 mm length error.

Our further work will focus on more detailed analysis of the shape of the Antarctic Krill as well as on approaches to integrate the Lateral and Dorsal views. In the future, we also aim to develop more maturity classifiers that operate on different hierarchies of maturity levels (e.g. J, F and M; J, MS, FS, MA and FA; and similar.). Most importantly, the Krill Tool will allow BAS researches to rapidly collect more Antarctic Krill data using our automated approaches. The forthcoming future data is expected to reinforce the shown models by increasing the number of the data-set samples. Future work will ensure making the data-set and code available to the public. 

\section{Acknowledgements}

This work was supported by the Engineering and Physical Sciences Research Council through the Centre for Doctoral Training in Agri-Food Robotics: AgriFoRwArdS [grant number EP/S023917/1].

\bibliography{bmvc_final}

\begin{thebibliography}{24}
\providecommand{\natexlab}[1]{#1}
\providecommand{\url}[1]{\texttt{#1}}
\expandafter\ifx\csname urlstyle\endcsname\relax
  \providecommand{\doi}[1]{doi: #1}\else
  \providecommand{\doi}{doi: \begingroup \urlstyle{rm}\Url}\fi

\bibitem[Arlot and Celisse(2010)]{Arlot_2010}
Sylvain Arlot and Alain Celisse.
\newblock A survey of cross-validation procedures for model selection.
\newblock \emph{Statistics Surveys}, 4\penalty0 (none), jan 2010.
\newblock \doi{10.1214/09-ss054}.
\newblock URL \url{https://doi.org/10.1214%2F09-ss054}.

\bibitem[Bezen et~al.(2020)Bezen, Edan, and Halachmi]{bezen2020computer}
Ran Bezen, Yael Edan, and Ilan Halachmi.
\newblock Computer vision system for measuring individual cow feed intake using rgb-d camera and deep learning algorithms.
\newblock \emph{Computers and electronics in agriculture}, 172:\penalty0 105345, 2020.

\bibitem[Borowiec et~al.(2022)Borowiec, Dikow, Frandsen, McKeeken, Valentini, and White]{borowiec2022deep}
Marek~L Borowiec, Rebecca~B Dikow, Paul~B Frandsen, Alexander McKeeken, Gabriele Valentini, and Alexander~E White.
\newblock Deep learning as a tool for ecology and evolution.
\newblock \emph{Methods in Ecology and Evolution}, 13\penalty0 (8):\penalty0 1640--1660, 2022.

\bibitem[Bowler et~al.(2020)Bowler, Fretwell, French, and Mackiewicz]{rs12122026}
Ellen Bowler, Peter~T. Fretwell, Geoffrey French, and Michal Mackiewicz.
\newblock Using deep learning to count albatrosses from space: Assessing results in light of ground truth uncertainty.
\newblock \emph{Remote Sensing}, 12\penalty0 (12), 2020.
\newblock ISSN 2072-4292.
\newblock \doi{10.3390/rs12122026}.
\newblock URL \url{https://www.mdpi.com/2072-4292/12/12/2026}.

\bibitem[Buda et~al.(2018)Buda, Maki, and Mazurowski]{buda2018systematic}
Mateusz Buda, Atsuto Maki, and Maciej~A Mazurowski.
\newblock A systematic study of the class imbalance problem in convolutional neural networks.
\newblock \emph{Neural networks}, 106:\penalty0 249--259, 2018.

\bibitem[{Django Software Foundation}()]{django}
{Django Software Foundation}.
\newblock Django.
\newblock URL \url{https://djangoproject.com}.

\bibitem[Dutta and Zisserman(2019)]{dutta2016via}
Abhishek Dutta and Andrew Zisserman.
\newblock The {VIA} annotation software for images, audio and video.
\newblock In \emph{Proceedings of the 27th {ACM} International Conference on Multimedia}. {ACM}, oct 2019.
\newblock \doi{10.1145/3343031.3350535}.
\newblock URL \url{https://doi.org/10.1145%2F3343031.3350535}.

\bibitem[Fan et~al.(2020)Fan, Li, Zhang, Tian, Wang, He, Zhang, and Huang]{fan2020line}
Shuxiang Fan, Jiangbo Li, Yunhe Zhang, Xi~Tian, Qingyan Wang, Xin He, Chi Zhang, and Wenqian Huang.
\newblock On line detection of defective apples using computer vision system combined with deep learning methods.
\newblock \emph{Journal of Food Engineering}, 286:\penalty0 110102, 2020.

\bibitem[French et~al.(2020)French, Mackiewicz, Fisher, Holah, Kilburn, Campbell, and Needle]{french2020deep}
Geoff French, Michal Mackiewicz, Mark Fisher, Helen Holah, Rachel Kilburn, Neil Campbell, and Coby Needle.
\newblock Deep neural networks for analysis of fisheries surveillance video and automated monitoring of fish discards.
\newblock \emph{ICES Journal of Marine Science}, 77\penalty0 (4):\penalty0 1340--1353, 2020.

\bibitem[He et~al.(2016)He, Zhang, Ren, and Sun]{he2016deep}
Kaiming He, Xiangyu Zhang, Shaoqing Ren, and Jian Sun.
\newblock Deep residual learning for image recognition.
\newblock In \emph{Proceedings of the IEEE conference on computer vision and pattern recognition}, pages 770--778, 2016.

\bibitem[Hodson(2022)]{hodson2022root}
Timothy~O Hodson.
\newblock Root-mean-square error (rmse) or mean absolute error (mae): When to use them or not.
\newblock \emph{Geoscientific Model Development}, 15\penalty0 (14):\penalty0 5481--5487, 2022.

\bibitem[Jacquet et~al.(2010)Jacquet, Pauly, Ainley, Holt, Dayton, and Jackson]{jacquet2010seafood}
Jennifer Jacquet, Daniel Pauly, David Ainley, Sidney Holt, Paul Dayton, and Jeremy Jackson.
\newblock Seafood stewardship in crisis.
\newblock \emph{Nature}, 467\penalty0 (7311):\penalty0 28--29, 2010.

\bibitem[Lin et~al.(2016)Lin, Doll{\'{a}}r, Girshick, He, Hariharan, and Belongie]{rick}
Tsung{-}Yi Lin, Piotr Doll{\'{a}}r, Ross~B. Girshick, Kaiming He, Bharath Hariharan, and Serge~J. Belongie.
\newblock Feature pyramid networks for object detection.
\newblock \emph{CoRR}, abs/1612.03144, 2016.
\newblock URL \url{http://arxiv.org/abs/1612.03144}.

\bibitem[Makarov and Denys(1981)]{makarov1981stages}
RR~Makarov and CJI Denys.
\newblock Stages of sexual maturity of euphausia suberba. biomass handbook 1.1, 1981.

\bibitem[Morris et~al.(1988)Morris, Watkins, Ricketts, Buchholz, and Priddle]{morris1988assessment}
DJ~Morris, Jonathan~L Watkins, CF~Ricketts, Friedrich Buchholz, and Julian Priddle.
\newblock An assessment of the merits of length and weight measurements of antarctic krill euphausia superba.
\newblock \emph{British Antarctic Survey Bulletin}, \penalty0 (79):\penalty0 27--50, 1988.

\bibitem[M{\"u}ller et~al.(1995)M{\"u}ller, Reinhardt, and Strickland]{muller1995neural}
Berndt M{\"u}ller, Joachim Reinhardt, and Michael~T Strickland.
\newblock \emph{Neural networks: an introduction}.
\newblock Springer Science \& Business Media, 1995.

\bibitem[Quintana et~al.(2015)Quintana, Torres, and Men{\'e}ndez]{quintana2015simplified}
Marcos Quintana, Juan Torres, and Jos{\'e}~Manuel Men{\'e}ndez.
\newblock A simplified computer vision system for road surface inspection and maintenance.
\newblock \emph{IEEE Transactions on Intelligent Transportation Systems}, 17\penalty0 (3):\penalty0 608--619, 2015.

\bibitem[Redmon et~al.(2016)Redmon, Divvala, Girshick, and Farhadi]{redmon2016you}
Joseph Redmon, Santosh Divvala, Ross Girshick, and Ali Farhadi.
\newblock You only look once: Unified, real-time object detection.
\newblock In \emph{Proceedings of the IEEE conference on computer vision and pattern recognition}, pages 779--788, 2016.

\bibitem[Schiermeier(2010)]{Schiermeier2010EcologistsFA}
Quirin Schiermeier.
\newblock Ecologists fear antarctic krill crisis.
\newblock \emph{Nature}, 467:\penalty0 15--15, 2010.

\bibitem[Szegedy et~al.(2013)Szegedy, Toshev, and Erhan]{szegedy2013deep}
Christian Szegedy, Alexander Toshev, and Dumitru Erhan.
\newblock Deep neural networks for object detection.
\newblock \emph{Advances in neural information processing systems}, 26, 2013.

\bibitem[Tian et~al.(2021)Tian, Shen, Wang, and Chen]{tian2021boxinst}
Zhi Tian, Chunhua Shen, Xinlong Wang, and Hao Chen.
\newblock Boxinst: High-performance instance segmentation with box annotations.
\newblock In \emph{Proceedings of the IEEE/CVF Conference on Computer Vision and Pattern Recognition}, pages 5443--5452, 2021.

\bibitem[V{\'e}lez-Rivera et~al.(2014)V{\'e}lez-Rivera, Blasco, Chanona-P{\'e}rez, Calder{\'o}n-Dom{\'\i}nguez, de~Jes{\'u}s Perea-Flores, Arzate-V{\'a}zquez, Cubero, and Farrera-Rebollo]{velez2014computer}
Nayeli V{\'e}lez-Rivera, Jos{\'e} Blasco, Jorge Chanona-P{\'e}rez, Georgina Calder{\'o}n-Dom{\'\i}nguez, Mar{\'\i}a de~Jes{\'u}s Perea-Flores, Israel Arzate-V{\'a}zquez, Sergio Cubero, and Reynold Farrera-Rebollo.
\newblock Computer vision system applied to classification of “manila” mangoes during ripening process.
\newblock \emph{Food and Bioprocess Technology}, 7:\penalty0 1183--1194, 2014.

\bibitem[Zhao et~al.(2019)Zhao, Zheng, Xu, and Wu]{zhao2019object}
Zhong-Qiu Zhao, Peng Zheng, Shou-tao Xu, and Xindong Wu.
\newblock Object detection with deep learning: A review.
\newblock \emph{IEEE transactions on neural networks and learning systems}, 30\penalty0 (11):\penalty0 3212--3232, 2019.

\bibitem[Zou et~al.(2023)Zou, Chen, Shi, Guo, and Ye]{zou2023object}
Zhengxia Zou, Keyan Chen, Zhenwei Shi, Yuhong Guo, and Jieping Ye.
\newblock Object detection in 20 years: A survey.
\newblock \emph{Proceedings of the IEEE}, 2023.

\end{thebibliography}
\end{document}